\documentclass{article}

\PassOptionsToPackage{numbers, compress, sort}{natbib}

\usepackage[preprint]{neurips_2025}

\usepackage[utf8]{inputenc} 
\usepackage[T1]{fontenc}    
\usepackage{hyperref}       
\usepackage{url}            
\usepackage{booktabs}       
\usepackage{amsfonts}       
\usepackage{nicefrac}       
\usepackage{microtype}      
\usepackage{xcolor}         

\usepackage{graphicx}
\usepackage{multirow}
\usepackage{subcaption}
\usepackage{enumitem}
\usepackage{makecell}
\usepackage{bm}
\usepackage{tabularx}

\title{SubGCache: Accelerating Graph-based RAG with Subgraph-level KV Cache}

\author{
\textbf{Qiuyu Zhu}$^{1}$ \quad
\textbf{Liang Zhang}$^{2}$\thanks{Corresponding author.} \quad
\textbf{Qianxiong Xu}$^{1}$ \quad
\textbf{Cheng Long}$^{1}$\footnotemark[1] \quad
\textbf{Jie Zhang}$^{1}$\\[0.4cm]
{\tt\{qiuyu002, liang012, qianxion001\}@e.ntu.edu.sg}\\
{\tt\{c.long, zhangj\}@ntu.edu.sg}\\[0.4cm]
$^{1}$Nanyang Technological University \\
$^{2}$Hong Kong University of Science and Technology (Guangzhou) \quad
}

\begin{document}

\maketitle


\begin{abstract}
Graph-based retrieval-augmented generation (RAG) enables large language models (LLMs) to incorporate structured knowledge via graph retrieval as contextual input, enhancing more accurate and context-aware reasoning. We observe that for different queries, it could retrieve similar subgraphs as prompts, and thus we propose SubGCache, which aims to reduce inference latency by reusing computation across queries with similar structural prompts ({\it i.e.}, subgraphs). Specifically, SubGCache clusters queries based on subgraph embeddings, constructs a representative subgraph for each cluster, and pre-computes the key-value (KV) cache of the representative subgraph. For each query with its retrieved subgraph within a cluster, it reuses the pre-computed KV cache of the representative subgraph of the cluster without computing the KV tensors again for saving computation. Experiments on two new datasets across multiple LLM backbones and graph-based RAG frameworks demonstrate that SubGCache consistently reduces inference latency with comparable and even improved generation quality, achieving up to 6.68$\times$ reduction in time-to-first-token (TTFT).

\end{abstract}

\section{Introduction}
\label{sec:intro}
\begin{figure}[t]
\centering\includegraphics[width=0.98\textwidth]{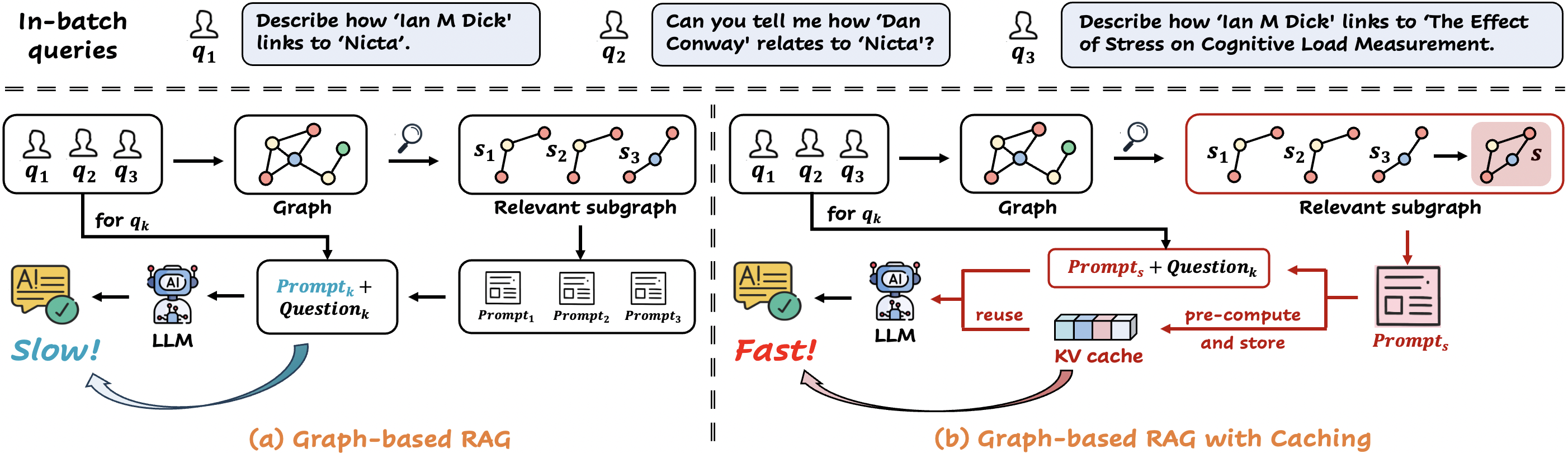}
  \caption{Overview of graph-based RAG without and with caching.}
  \label{fig:intro}
  \vspace{-3mm}
\end{figure}

Retrieval-augmented generation (RAG)~\cite{borgeaud2022improving,lewis2020retrieval,ram2023context,trivedi2022interleaving} enhances large language models (LLMs)~\cite{achiam2023gpt,chowdhery2023palm,grattafiori2024llama} by retrieving and integrating external knowledge based on text similarity, enabling more accurate and contextually enriched generation. Building on its success in language-focused tasks \cite{zhang2023prompting,zhang2024benchmarking}, recent efforts \cite{guo2024lightrag,he2024g,hu2024grag} have extended RAG to graph data~\cite{jin2024large,li2023survey,wang2023can}, giving rise to graph-based RAG~\cite{he2024g,hu2024grag}, which leverages textual graphs as external knowledge sources to help model entity relations across documents and support complex reasoning over structured knowledge. As illustrated in Figure~\ref{fig:intro}(a), upon receiving a user query $q_k$ and a textual graph $G$, graph-based RAG first retrieves the most relevant subgraph from $G$ and constructs a subgraph prompt. 
This prompt is then combined with the query to form an augmented input for the LLM to generate the final response.

While proven effective, existing graph-based RAG systems are primarily designed for single-query settings, where each query is processed independently by the LLM, as shown in Figure~\ref{fig:intro}(a). However, in many real-world scenarios \cite{ding2011batch,choudhury2018batch,zhang2024enhanced,bouros2024hint} such as medical question answering over biomedical knowledge graphs~\cite{han2024retrieval}, queries are batch-submitted, arrive in large volumes simultaneously, and are processed jointly, naturally forming in-batch workloads for graph-based RAG. Figure~\ref{fig:intro}(b) illustrates a typical in-batch scenario, where a group of queries $q_1$, $q_2$, and $q_3$ are submitted and processed together. Each query triggers the retrieval of a relevant subgraph from the external textual graph, resulting in subgraphs $s_1$, $s_2$, and $s_3$. 
In practice, these retrieved subgraphs may exhibit significant overlap. For instance, we can observe that $s_1$ and $s_2$ are identical, while $s_3$ shares large structural components with them. 
Despite such redundancy, existing methods process each query in isolation, repeatedly encoding and reasoning over the overlapping subgraph content, leading to unnecessary computation. These observations call for a rethinking of graph-based RAG in a new in-batch setting and raise a natural question: how can we effectively exploit structural redundancy across different queries to eliminate redundant computation and improve overall system efficiency?

An intuitive answer to this question is to introduce a caching mechanism that stores and reuses previously computed results from the LLM to avoid repeated computation. In fact, recent efforts~\cite{gim2024prompt,jin2024ragcache,zheng2023efficiently} have explored similar strategies in purely textual settings, where each cached unit corresponds to an independent sentence or document chunk. For instance, Prompt Cache~\cite{gim2024prompt} stores the pre-computed attention states of frequently occurring text segments, improving efficiency through inference-time reuse. However, these approaches are inherently limited to sequential text data and assume exact lexical repetition. They are not applicable to graph-based RAG, where redundancy manifests at the structural level, and each cached unit should be a structured subgraph composed of interconnected nodes and edges, with information organized topologically rather than sequentially. This structural nature of graph-based RAG introduces two critical and unique challenges: 

\begin{itemize}[leftmargin=*]
    \item \textbf{Challenge 1: Structural redundancy identification.} In-batch queries may retrieve subgraphs that are structurally and semantically similar, but such overlap is neither explicitly known beforehand nor easily detectable. Here, the key challenge lies in effectively comparing retrieved subgraphs, which may differ in node identifiers, local context, or graph topology, to determine whether meaningful overlap exists.
    \item \textbf{Challenge 2: Structural redundancy exploitation.} Even when overlap is correctly identified across queries, the retrieved subgraphs are generally partially shared. Unlike existing methods for sequential text \cite{gim2024prompt,jin2024ragcache,zheng2023efficiently}, which assume reuse over identical units, overlapping subgraphs may differ in size, topology, or node alignment. Here, another key challenge is to effectively reason over these partially shared structures across queries to reduce redundant computation, while still preserving the useful relational context necessary for accurate response generation.
\end{itemize}

To tackle these challenges, we propose SubGCache (\textbf{Subg}raph-level key-value \textbf{Cache}), a lightweight and efficient plug-and-play caching framework tailored for graph-based RAG under the in-batch query setting. It consists of two main components:
\begin{itemize}[leftmargin=*]
    \item \textbf{Design 1: Query clustering based on subgraph similarity.} SubGCache performs hierarchical clustering to in-batch queries based on the embeddings of their retrieved subgraphs, generated by the pretrained Graph Neural Network (GNN) encoder used in graph-based RAG. These embeddings encode both semantic and structural information, allowing the system to automatically identify subgraph-level redundancy across queries. Queries with highly overlapping subgraphs are then effectively grouped together for shared processing, thereby addressing the challenge of structural redundancy identification.

    \item \textbf{Design 2: Representative subgraph construction and subgraph-level cache reuse.} To facilitate effective reasoning over partially overlapping subgraphs while preserving the useful relational context, SubGCache introduces the concept of representative subgraph as shared structural input for each query cluster. Specifically, for each cluster, it merges the retrieved subgraphs from all queries within this cluster into a single representative subgraph that preserves the topology necessary for accurate response generation. To exploit this shared structure and eliminate redundant computation, the key-value (KV) cache mechanism is further introduced to pre-compute KV tensors of the representative subgraph and reuse them across all queries in the cluster. This cluster-wise strategy addresses the challenge of reusing partial structural overlaps by aligning similar subgraphs into a unified representation and caching its computation. As illustrated in Figure~\ref{fig:intro}(b), assume queries $q_1$, $q_2$, and $q_3$ are clustered together. SubGCache generates a representative subgraph $s$ by merging their retrieved subgraphs $s_1$, $s_2$, and $s_3$, constructs the corresponding prompt prefix $\textit{Prompt}_{s}$, and computes its KV tensors within the LLM, which are then stored in GPU memory. For each query $q_k \in \{q_1, q_2, q_3\}$, SubGCache directly appends the query-specific question tokens to the cached prefix, allowing the model to bypass recomputation of the shared subgraph context. By reusing the newly proposed subgraph-level KV cache across all queries in the cluster, SubGCache significantly reduces inference latency while maintaining strong generation quality.
\end{itemize}
 
Extensive experiments across two datasets and multiple LLM backbones validate the latency reduction and generation quality of SubGCache. Our main contributions are summarized as follows:

\begin{itemize}[leftmargin=*]
    \item \textbf{Conceptually:} We formulate a new research problem under the in-batch query setting, aiming to accelerate graph-based RAG via batch-level processing. To the best of our knowledge, this is the first work to accelerate graph-based RAG and explore batch-level execution in this context.    
    \item \textbf{Methodologically:} We propose SubGCache, a lightweight and plug-and-play framework for subgraph-level prompt caching that addresses the unique challenges of structural redundancy identification and exploitation in retrieved subgraphs. It is simple to implement, and both highly effective and efficient in practice. Notably, this is also the first attempt to introduce prompt caching into graph-based RAG.

    \item \textbf{Empirically:} Experiments on two datasets across multiple LLM backbones and graph-based RAG frameworks demonstrate that SubGCache consistently reduces inference latency while maintaining or even enhancing generation quality. For example, with Llama-3.2-3B, it achieves up to 5.69$\times$ speedup with 2.00\% accuracy gain on the Scene Graph, and 6.52$\times$ speedup with 1.00\% accuracy gain on the OAG dataset.
    
\end{itemize}  
\section{Related Work}
\label{sec:related_work}
\noindent \textbf{RAG.} 
RAG~\cite{fan2024survey,hu2024rag,huang2025survey,lewis2020retrieval,ram2023context,trivedi2022interleaving,yu2024evaluation,zhao2024retrieval} enhances LLMs by retrieving external knowledge to mitigate hallucination~\cite{huang2025survey} and improve reliability~\cite{gao2023retrieval}. Recently, graph-based RAG was proposed~\cite{guo2024lightrag,he2024g,hu2024grag}, which retrieves query-relevant subgraphs from textual graphs and performs generation by jointly leveraging text and structures. For example, G-Retriever~\cite{he2024g} retrieves individual nodes and edges and reconstructs query-specific subgraphs for generation, while GRAG~\cite{hu2024grag} retrieves subgraphs directly by embedding $k$-hop ego networks and pruning irrelevant components. These graph-based RAG methods focus primarily on single-query processing and overlook the holistic optimization opportunities enabled by in-batch query execution. Moreover, they pay little attention to inference efficiency, concentrating solely on improving retrieval and generation quality. In this paper, we aim to improve the inference efficiency of graph-based RAG by exploiting structural redundancy through batch-level processing.

\noindent \textbf{KV cache reuse.} Recent efforts~\cite{gim2024prompt,jin2024ragcache,zheng2023efficiently,yao2025cacheblend,jin2024compute,lisharedcontextbench,lu2024turborag} have explored reusing KV cache to reduce redundant computation during LLM inference, primarily within text-based scenarios. For instance, SGLang~\cite{zheng2023efficiently} identifies reusable intermediate states across different requests in multi-turn conversations, while Prompt Cache~\cite{gim2024prompt} enables flexible token reuse by ensuring each prompt module is self-contained and semantically independent. Furthermore, RAGCache~\cite{jin2024ragcache} exploits the retrieved document sequences to construct a multilevel caching system, improving efficiency without altering generation outputs. However, these approaches are tailored to text-only settings and do not address the unique challenges associated with graph retrieval, where the retrieved subgraphs are inherently interconnected and leveraging their topological structure is critical to maintain generation quality. To bridge this gap, we introduce a novel caching paradigm based on structured subgraphs and propose SubGCache, a lightweight and efficient framework for subgraph-level prompt caching that identifies and exploits the structural redundancy in retrieved subgraphs.

\section{Methodology}
\label{sec:methodology}

\begin{figure}[t]
\centering\includegraphics[width=1.0\textwidth]{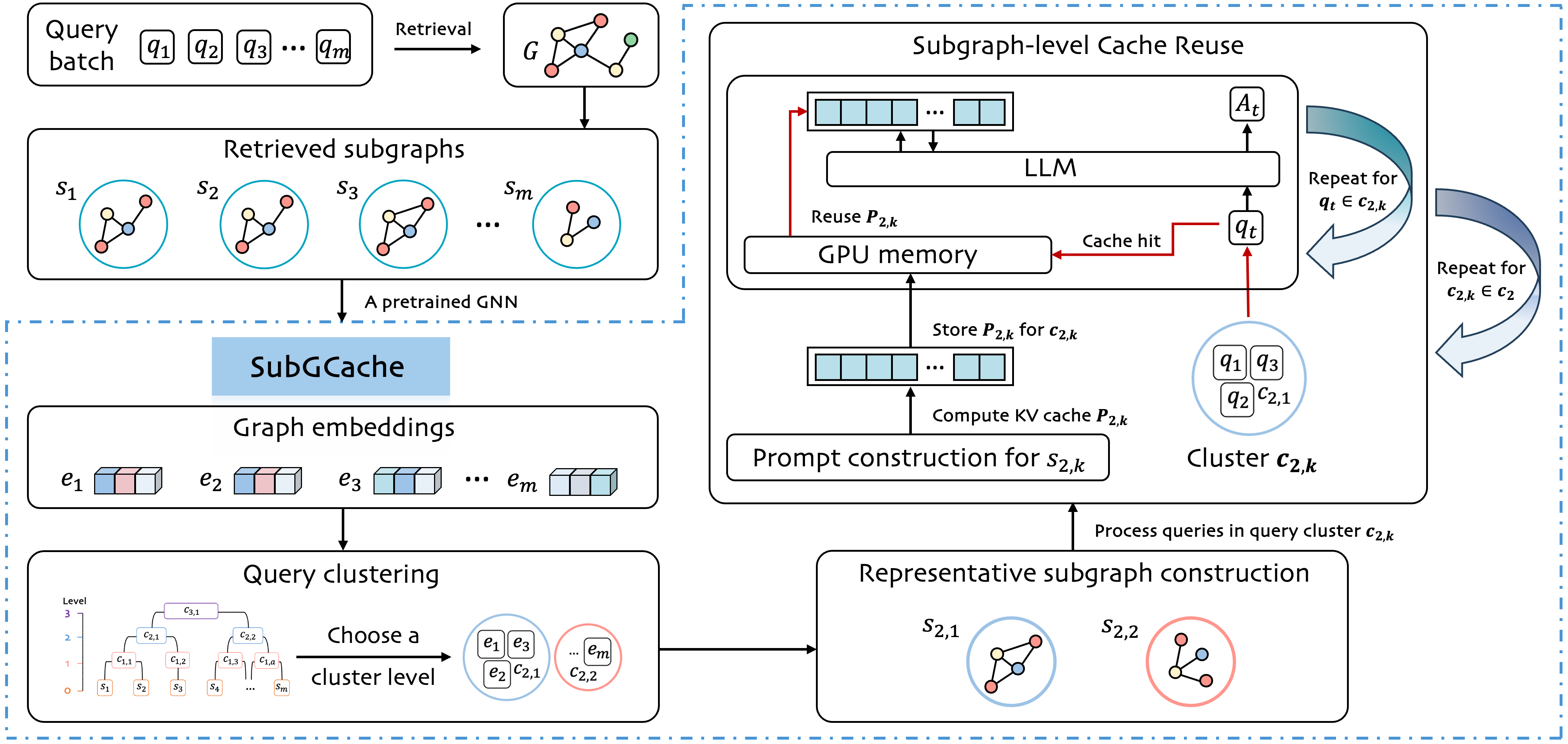}
  \caption{Overview of SubGCache and its integration into the standard graph-based RAG pipeline.}
  \label{fig:model}
  \vspace{-2mm}
\end{figure}

We consider a new in-batch setting for graph-based RAG, where a batch of queries $\{q_1, q_2, \dots, q_m\}$ is issued simultaneously to a shared system. In the standard graph-based RAG pipeline, each query $q_i$ retrieves a corresponding subgraph $s_i$ from a textual graph $G$, and then an LLM generates a response $a_i$ based on the augmented input formed by $q_i$ and $s_i$.

While effective, this per-query processing paradigm results in substantial redundant computation. To address this limitation, we propose SubGCache, a lightweight and efficient plug-and-play caching framework that identifies shared subgraphs across queries and eliminates redundant computation by caching and reusing their KV tensors. The overall design and workflow are shown below.

\subsection{Architecture Overview}
\label{sec:overview}
Figure~\ref{fig:model} provides an overview of SubGCache and its integration into the standard graph-based RAG pipeline. Specifically, given a textual graph $G$,  a batch of queries $\{q_1,q_2,\dots,q_m\}$, and their corresponding retrieved subgraphs $\{s_1,s_2,\dots,s_m\}$, SubGCache is designed to reduce redundant computation by leveraging structural redundancy across queries through the following three key steps: (1) Query clustering: In-batch queries are grouped based on structural and semantic similarities in their retrieved subgraphs, enabling the identification of shared subgraph components. (2) Representative subgraph construction: For each cluster, we merge the nodes and edges of all associated subgraphs to create a representative subgraph that preserves the relational context required for high-quality response generation. (3) Subgraph-level cache reuse: SubGCache processes queries in a cluster-wise manner. For each cluster, it computes the KV cache for the representative subgraph only once, reuses it across all associated queries, and releases it before moving to the next. This substantially reduces redundant computation and improves inference efficiency, without compromising generation quality.

\subsection{Query Clustering}
\noindent \textbf{Graph Embedding via Pretrained GNN.} 
The key intuition behind query clustering is that in-batch queries may retrieve subgraphs that are structurally and semantically similar. However, such overlap is neither known beforehand nor trivial to detect, as retrieved subgraphs may differ in node identifiers, local context, or overall topology. To address this challenge, we encode each retrieved subgraph into a graph embedding using a pretrained GNN initialized with SentenceBERT-based node features—the same setup used for soft prompt construction in existing graph-based RAG. These embeddings capture both semantic and structural characteristics, enabling effective comparison across subgraphs.

\noindent \textbf{Hierarchical Clustering.} Once the subgraph embeddings $\{e_1,e_2,\dots,e_m\}$ are obtained, we perform hierarchical clustering over these embeddings to group similar subgraphs. As a result, subgraphs with substantial overlap (\textit{\textit{i.e.}}, subgraph-level redundancy across queries) are automatically assigned to the same cluster. Their corresponding queries are thus grouped for shared processing, effectively addressing the challenge of structural redundancy identification.

\noindent \textbf{Example.} As illustrated in Figure~\ref{fig:model}, given a batch of queries $\{q_1,q_2,\dots,q_m\}$ and their corresponding retrieved subgraphs $\{s_1,s_2,\dots,s_m\}$, we first encode each subgraph into an embedding $\{e_1,e_2,\dots,e_m\}$ using the pretrained GNN. Hierarchical clustering is then applied with a predefined number of clusters (\textit{i.e.}, $c=2$) to group similar embeddings together. For instance, embeddings $e_1$, $e_2$ and $e_3$ are assigned to cluster $C_{2,1}$, while the remaining form cluster $C_{2,2}$. Consequently, both the retrieved subgraphs and their associated queries are grouped accordingly, laying the foundation for downstream subgraph-level cache reuse.

\subsection{Representative Subgraph Construction}
Although queries with significant structural redundancy can be effectively identified through GNN-based clustering, the retrieved subgraphs are generally partially shared, as they may differ in size, topology, or node alignment. This contrasts with text-based reuse methods, where cached units such as sentences or document chunks 
are typically assumed to be identical and easily shareable. 

To address this challenge, we introduce a simple and effective representative subgraph as the shared structural input and natural cached unit for each query cluster. It is constructed by taking the union of all nodes and edges from the subgraphs retrieved by the queries within a specific cluster. The resulting structure captures the full relational context shared across the cluster and serves as a comprehensive, reusable input that supports both accurate response generation and structural redundancy elimination.

\noindent \textbf{Example.} As presented in Figure~\ref{fig:model}, suppose the subgraph embeddings are grouped into two clusters: $C_{2,1}$ containing $s_1$, $s_2$, and $s_3$, and $C_{2,2}$ containing the remaining subgraphs. For cluster $C_{2,1}$, we construct the representative subgraph $s_{2,1}$ by merging all nodes and edges from the corresponding retrieved subgraphs $\{s_1,s_2,s_3\}$. Likewise, another representative subgraph $s_{2,2}$ is constructed for $C_{2,2}$ by merging the subgraphs assigned to that cluster.

\subsection{Subgraph-level Cache Reuse}
While representative subgraphs provide unified structural input for each query cluster, efficiently leveraging them during response generation remains non-trivial. To achieve this, SubGCache adopts a cluster-wise processing strategy, enabled by a subgraph-level caching mechanism that pre-computes attention states once and reuses them across all queries within the same cluster. Specifically, for each cluster, SubGCache constructs a prompt based on its representative subgraph, following standard graph-based RAG pipelines. The prompt is then fed into the LLM to pre-compute intermediate attention states across transformer layers, which are stored in GPU memory as a cluster-wise KV cache and reused by all queries in the cluster. When processing each query, SubGCache appends query-specific question tokens to the cached subgraph prompt, enabling the model to directly leverage the shared structural context without redundant computation. 

Once all queries in a cluster are processed, the corresponding KV cache is released to free GPU memory before moving to the next. This cluster-wise cache management eliminates redundant computation, reduces memory usage, and ensures scalability for large in-batch query workloads.

\noindent \textbf{Example.} Continuing the example in Figure~\ref{fig:model}, after obtaining clusters $C_{2,1}$ and $C_{2,2}$ with their representative subgraphs $s_{2,1}$ and $s_{2,2}$, SubGCache processes them sequentially. Specifically, it first processes $C_{2,1}$ by constructing a prompt for $s_{2,1}$ and computing its KV cache $P_{2,1}$, which is then stored in GPU memory.  All queries in $C_{2,1}$ achieve cache hits by reusing this shared KV cache. Once all queries in $C_{2,1}$ are served, the cache is released to free GPU memory. Then, SubGCache repeats the procedure for $C_{2,2}$ using $s_{2,2}$ and $P_{2,2}$. This cluster-wise reuse and release strategy ensures efficient memory usage and eliminates redundant computation, even with large in-batch workloads.

\noindent \textbf{Discussion.} SubGCache enables flexible control over cache reuse granularity by adjusting the clustering level. Finer clustering (\textit{i.e.}, more clusters) yields more query-specific prompts, but limits reuse opportunities. In contrast, coarser clustering (\textit{i.e.}, fewer clusters) promotes greater reuse by grouping more queries together and generating subgraphs with broader context. This often enhances generation quality, although it may also introduce minor noise in rare cases, as observed in our experiments. Notably, when each query forms its own cluster, the method naturally reduces to standard graph-based RAG.
  
\section{Experiments}
\label{sec:exp} 

\begin{table}[t]
  \caption{Dataset statistics.}
  \label{tab:data}
  \centering
  \setlength{\tabcolsep}{6pt}
  \resizebox{1.00\textwidth}{!}{
  \begin{tabular}{cccccc}
    \toprule
    Dataset & \#Nodes & \#Relations & \#Queries & Node Attribute & Edge Attribute \\
    \midrule
    Scene Graph & 22 & 147 & 426 & Entity attributes (\textit{e.g.}, color) & Relations (\textit{e.g.}, spatial relations) \\
    OAG & 1071 & 2022 & 3434 & Entity name & Relations (\textit{e.g.}, predicates)
    \\
    \bottomrule
  \end{tabular}
  }
  \vspace{-3mm}
\end{table}

\subsection{Experimental Setup}
\label{sec:experimental_setup}

\noindent\textbf{Datasets:} We construct two new datasets, Scene Graph and OAG, to support in-batch query evaluation for graph-based RAG. Key statistics are summarized in Table~\ref{tab:data}, with details in Appendix~\ref{sec:data_detail}.

\noindent\textbf{Setup:} 
We adopt two representative graph-based RAG methods, G-Retriever~\cite{he2024g} and GRAG~\cite{hu2024grag}, as our baseline models. SubGCache is then integrated as a plug-and-play module, resulting in G-Retriever+SubGCache and GRAG+SubGCache. All methods are tested with different LLM backbones: Llama-3.2-3B~\cite{grattafiori2024llama}, Llama-2-7B~\cite{touvron2023llama}, Mistral-7B~\cite{mosaicml2023introducing}, and Falcon-7B~\cite{penedo2023refinedweb}. All experiments are conducted in an inference-only setting with frozen LLMs. We evaluate performance with four metrics: accuracy (ACC), response time (RT), time-to-first-token (TTFT), and prefill and first token time (PFTT). ACC is reported as a percentage (\%), and the other metrics in milliseconds (ms). Configuration and metric details are in Appendix~\ref{sec:config} and~\ref{sec:metrics}, respectively.

\begin{table}
  \caption{Overall performance. The best results are highlighted in bold.}
  \label{tab:main}
  \centering
  \setlength{\tabcolsep}{8pt}
  \resizebox{1.0\textwidth}{!}{
    \begin{tabular}{l|cccc|cccc}
      \toprule
      \multirow{2}{*}{Model} & \multicolumn{4}{c|}{Scene Graph} & \multicolumn{4}{c}{OAG} \\
      & ACC$\uparrow$ & RT$\downarrow$ & TTFT$\downarrow$ & PFTT$\downarrow$ & ACC$\uparrow$ & RT$\downarrow$ & TTFT$\downarrow$ & PFTT$\downarrow$ \\
      \midrule
      \addlinespace[1.0ex]
      \multicolumn{9}{c}{\textbf{Backbone: Llama-3.2-3B}} \\
      \midrule
      G-Retriever & 62.00 & 664.71 & 642.86 & 321.26 & 96.00 & 974.94 & 921.00 & 245.07 \\
      G-Retriever+SubGCache & \textbf{64.00} & \textbf{132.93} & \textbf{112.93} & \textbf{26.92} & \textbf{97.00} & \textbf{190.73} & \textbf{141.19} & \textbf{29.94} \\ 
     $\Delta_{G-Retriever}$ &$\uparrow$ 2.00 &$\uparrow$ 5.00$\times$ &$\uparrow$ 5.69$\times$ &$\uparrow$ 11.93$\times$ &$\uparrow$ 1.00 &$\uparrow$ 5.11$\times$ &$\uparrow$ 6.52$\times$ &$\uparrow$ 8.19$\times$ \\
      \midrule
      GRAG & 60.00 & 559.17 & 540.99 & 400.18 & \textbf{98.00} & 243.50 & 186.61 & 82.63 \\
      GRAG+SubGCache & \textbf{61.00} & \textbf{154.79} & \textbf{132.60} & \textbf{19.77} & 97.00 & \textbf{174.79} & \textbf{124.44} & \textbf{30.84} \\ 
     $\Delta_{GRAG}$ &$\uparrow$ 1.00 &$\uparrow$ 3.61$\times$ &$\uparrow$ 4.08$\times$ &$\uparrow$ 19.77$\times$ &$\downarrow$ 1.00 &$\uparrow$ 1.39$\times$ &$\uparrow$ 1.50$\times$ &$\uparrow$ 2.68$\times$ \\ 
      \midrule
      \addlinespace[1.0ex]
      \multicolumn{9}{c}{\textbf{Backbone: Llama-2-7B}} \\
      \midrule
      G-Retriever & 59.00  & 970.04 & 938.44 & 705.51 & \textbf{94.00} & 922.83 & 852.95 & 524.32 \\
      G-Retriever+SubGCache & \textbf{66.00} & \textbf{168.52} & \textbf{140.54} & \textbf{45.55} & \textbf{94.00} & \textbf{282.52} & \textbf{217.26} & \textbf{60.63} \\ 
     $\Delta_{G-Retriever}$ &$\uparrow$ 7.00 & $\uparrow$ 5.76$\times$ &$\uparrow$ 6.68$\times$ &$\uparrow$ 15.49$\times$ & 0.00 &$\uparrow$ 3.27$\times$ &$\uparrow$ 3.93$\times$ &$\uparrow$ 8.65$\times$ \\ 
      \midrule
      GRAG & 56.00 & 1299.79 & 1264.70 & 924.11 & \textbf{99.00} & 441.97 & 375.13 & 217.17 \\
      GRAG+SubGCache & \textbf{57.00} & \textbf{234.87} & \textbf{202.96} & \textbf{50.53} & \textbf{99.00} & \textbf{258.67} & \textbf{188.84} & \textbf{62.23} \\ 
     $\Delta_{GRAG}$ &$\uparrow$ 1.00 &$\uparrow$ 5.53$\times$ &$\uparrow$ 6.23$\times$ &$\uparrow$ 18.29$\times$ & 0.00 &$\uparrow$ 1.71$\times$ &$\uparrow$ 1.99$\times$ &$\uparrow$ 3.49$\times$ \\ 
      \midrule
      \addlinespace[1.0ex]
      \multicolumn{9}{c}{\textbf{Backbone: Mistral-7B}} \\
      \midrule
      G-Retriever & \textbf{66.00} & 960.42 & 930.76 & 742.55 & \textbf{99.00} & 766.29 & 687.10 & 552.65 \\
      G-Retriever+SubGCache & \textbf{66.00} & \textbf{236.21} & \textbf{204.32} & \textbf{52.11} & \textbf{99.00} & \textbf{315.35} & \textbf{237.74} & \textbf{63.42} \\ 
     $\Delta_{G-Retriever}$ & 0.00 &$\uparrow$ 4.07$\times$ &$\uparrow$ 4.56$\times$ &$\uparrow$ 14.25$\times$ & 0.00 &$\uparrow$ 2.43$\times$ &$\uparrow$ 2.89$\times$ &$\uparrow$ 8.71$\times$ \\ 
      \midrule
      GRAG & 57.00 & 1113.75 & 1081.97 & 966.54 & \textbf{99.00} & 539.39 & 458.70 & 243.82 \\
      GRAG+SubGCache & \textbf{66.00} & \textbf{194.68} & \textbf{164.01} & \textbf{52.44} & \textbf{99.00} & \textbf{237.04} & \textbf{159.25} & \textbf{63.04} \\ 
     $\Delta_{GRAG}$ &$\uparrow$ 9.00 &$\uparrow$ 5.72$\times$ &$\uparrow$ 6.60$\times$ &$\uparrow$ 18.43$\times$ & 0.00 &$\uparrow$ 2.28$\times$ &$\uparrow$ 2.88$\times$ &$\uparrow$ 3.87$\times$ \\
      \midrule
      \addlinespace[1.0ex]
      \multicolumn{9}{c}{\textbf{Backbone: Falcon-7B}} \\
      \midrule
      G-Retriever & 64.00  & 826.56 & 790.46 & 702.11 & \textbf{98.00} & 1049.20 & 964.67 & 526.74 \\
      G-Retriever+SubGCache & \textbf{66.00} & \textbf{195.29} & \textbf{159.81} & \textbf{52.16} & 97.00  & \textbf{374.53} & \textbf{294.55} & \textbf{59.66} \\ 
     $\Delta_{G-Retriever}$ &$\uparrow$ 2.00 &$\uparrow$ 4.23$\times$ &$\uparrow$ 4.95$\times$ &$\uparrow$ 13.46$\times$ &$\downarrow$ 1.00 &$\uparrow$ 2.80$\times$ &$\uparrow$ 3.28$\times$ &$\uparrow$ 8.83$\times$ \\
      \midrule
      GRAG & 57.00  & 1142.68 & 1105.78 & 954.17 & \textbf{97.00} & 483.21 & 400.54 & 198.88 \\
      GRAG+SubGCache & \textbf{60.00} & \textbf{272.45} & \textbf{238.04} & \textbf{50.49} & 96.00 & \textbf{249.28} & \textbf{169.03} & \textbf{59.18} \\ 
     $\Delta_{GRAG}$ &$\uparrow$ 3.00 &$\uparrow$ 4.19$\times$ &$\uparrow$ 4.65$\times$ &$\uparrow$ 18.90$\times$ &$\downarrow$ 1.00 &$\uparrow$ 1.94$\times$ &$\uparrow$ 2.37$\times$ & $\uparrow$ 3.36$\times$ \\
      \bottomrule
    \end{tabular}
  }
  \vspace{-3mm}
\end{table}

\subsection{Main Results}
\label{sec:overall_performance}
Table~\ref{tab:main} summarizes the overall results on both datasets using four LLM backbones.

\noindent \textbf{Reduced latency with comparable effectiveness.} Compared to the baseline models G-Retriever and GRAG, integrating our SubGCache framework ({\it i.e.}, G-Retriever+SubGCache and GRAG+SubGCache) consistently reduces latency across both datasets. Specifically, for G-Retriever, SubGCache achieves up to 5.76$\times$ / 5.11$\times$ speedup in RT, 6.68$\times$ / 6.52$\times$ in TTFT, and 15.49$\times$ / 8.19$\times$ reduction in PFTT on the Scene Graph and OAG datasets, respectively. For GRAG, it yields 5.72$\times$ / 2.28$\times$ in RT, 6.60$\times$ / 2.88$\times$ in TTFT, and 18.43$\times$ / 3.87$\times$ in PFTT. These substantial latency reductions come with comparable or even improved accuracy: up to 9.00\% gain on Scene Graph, and only a minor drop ({\it i.e.}, 1.00\%) in rare cases on OAG dataset.

\noindent \textbf{Consistent improvement across LLM backbones.} SubGCache consistently reduces latency with comparable generation quality across different LLM backbones, regardless of architectural or scale differences. This confirms its robustness and generalization as a plug-and-play optimization.

\noindent \textbf{Understanding why SubGCache works.} SubGCache significantly reduces inference latency with comparable accuracy by addressing two key challenges in graph-based RAG: identifying and exploiting structural redundancy. (1) It clusters in-batch queries based on the semantic and structural similarity of their retrieved subgraphs using pretrained GNN embeddings, enabling queries with overlapping context to be grouped and processed together. (2) For each cluster, it constructs a representative subgraph by merging the retrieved subgraphs into a unified structure. The KV cache for this shared input is computed once and reused across all queries in the cluster, avoiding redundant computation while preserving relational context. In rare cases, the merged context may introduce minor noise, leading to slight degradation in effectiveness. Together, these two designs explain the observed latency reduction and stable generation quality across datasets and LLM backbones, highlighting SubGCache’s practical value as an efficient caching strategy for graph-based RAG.

\begin{figure}
\centering\includegraphics[width=0.98\textwidth]{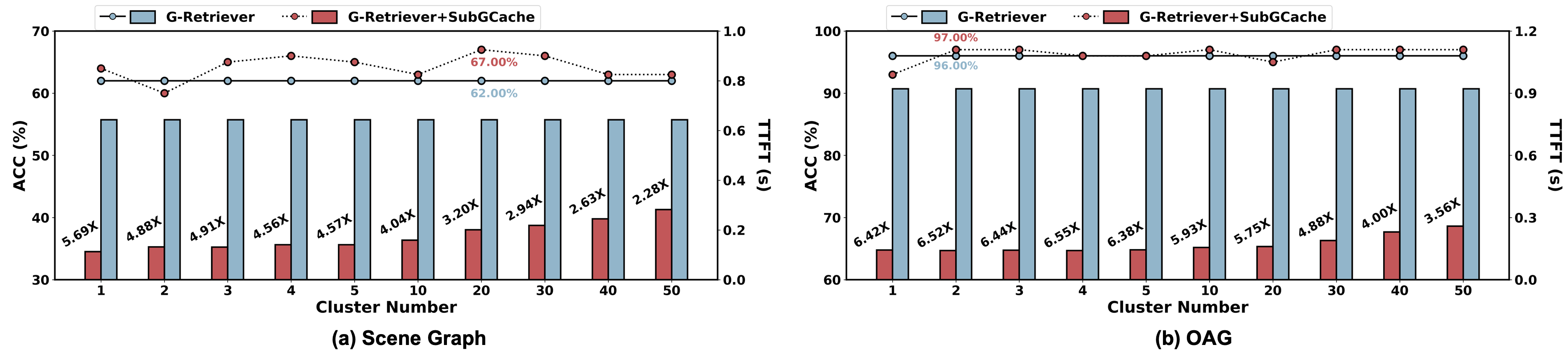}
  \caption{Impact of cluster number on ACC (\%) and TTFT (s).}
  \label{fig:cluster}
  \vspace{-3mm}
\end{figure}

\subsection{Impact of Cluster Number}
\label{sec:cluster}
To evaluate the effect of cluster number, we compare G-Retriever with G-Retriever+SubGCache by varying cluster numbers in \{1, 2, 3, 4, 5, 10, 20, 30, 40, 50\}, and report performance on both datasets using the Llama-3.2-3B backbone, as shown in Figure~\ref{fig:cluster}.

\noindent \textbf{Trade-off between latency and accuracy.} As observed, finer clustering (\textit{i.e.}, more clusters) tends to preserve more query-specific context, which can improve accuracy, while coarser clustering boosts cache reuse and reduces latency. However, this trade-off is not strictly monotonic. Both latency and accuracy fluctuate across cluster settings due to competing factors. Fewer clusters enable more frequent reuse but lead to larger representative subgraphs, increasing prompt length and cache overhead. More clusters reduce reuse opportunities but produce shorter prompts. This results in a non-linear latency trend, where TTFT does not steadily increase with cluster number. On the accuracy side, coarser clustering may improve quality by aggregating richer subgraph context, or slightly degrades performance by introducing irrelevant information. 

Despite these variations, SubGCache performs well even with small cluster number. On Scene Graph, the 1-cluster setting achieves 5.69$\times$ speedup in TTFT while surpassing baseline's accuracy. On OAG, the 2-cluster setting  yields a favorable result, achieving a 1.00\% accuracy gain alongside 6.52$\times$ speedup in TTFT, respectively. These results highlight the importance of selecting an appropriate clustering granularity to balance latency and accuracy.

\begin{figure}
\centering\includegraphics[width=0.95\textwidth]{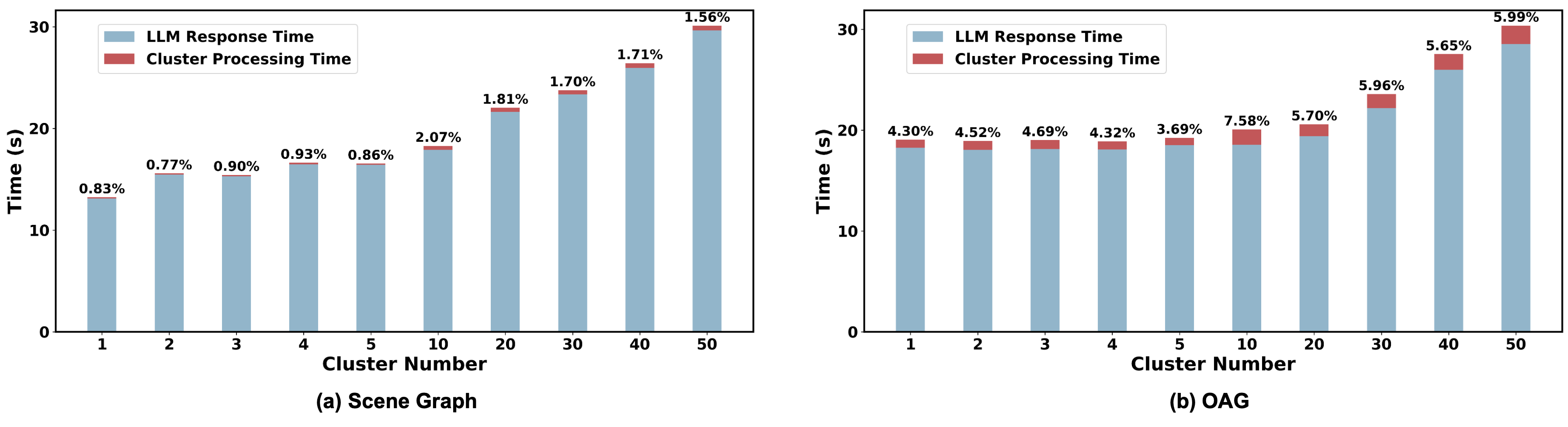}
  \caption{Cluster processing time vs. LLM response time by varying cluster numbers.}
  \label{fig:percent}
  \vspace{-3mm}
\end{figure}

\subsection{Cluster Processing Time} 
\label{sec:cluster_process}
Figure~\ref{fig:percent} compares the LLM response time (blue) and cluster processing time (red) of G-Retriever+SubGCache under varying cluster numbers on both datasets. Cluster processing time includes graph encoding, hierarchical clustering, and representative subgraph construction. We summarize four key observations: 

\noindent \textbf{Minimal processing overhead.}
Cluster processing time remains low across all cluster configurations. On Scene Graph, it accounts for less than 2.1\% of total latency, and below 6\% even on the larger OAG dataset with 50 clusters. These results show that SubGCache’s clustering stage introduces only modest overhead relative to total inference time.

\noindent \textbf{Higher cost on larger graphs.}  
OAG incurs higher processing time than Scene Graph, primarily due to its larger graph size. These properties result in larger retrieved subgraphs, increasing the number of nodes and edges to encode, and leading to higher computational cost during both GNN-based embedding and representative subgraph construction. 

\noindent \textbf{Non-monotonic variation.} Cluster processing time does not increase linearly with cluster number. While more clusters require more representative subgraphs, each individual cluster is smaller, reducing per-cluster encoding time. Additionally, hierarchical clustering complexity depends on the number of inputs, rather than the number of output clusters, contributing to the non-linear trend.

\noindent \textbf{LLM response time generally increases with cluster number.} Finer clustering limits cache reuse across queries, leading to longer response times. Slight fluctuations arise from larger merged subgraphs, which generate longer prompts and incur higher inference costs.

\begin{table}
  \caption{Impact of different linkage strategies.}
  \label{tab:linkage}
  \centering
  \setlength{\tabcolsep}{8pt}
  \resizebox{1.0\textwidth}{!}{
    \begin{tabular}{c|c|cccc|cccc}
      \toprule
      \multirow{2}{*}{} & \multirow{2}{*}{Strategies} & \multicolumn{4}{c|}{Scene Graph} & \multicolumn{4}{c}{OAG} \\
      & &$\Delta$ACC &$\Delta$RT &$\Delta$TTFT &$\Delta$PFTT &$\Delta$ACC &$\Delta$RT &$\Delta$TTFT &$\Delta$PFTT \\
      \midrule
      \multirow{5}{*}{$\Delta_{G-Retriever}$}
      & Ward  & \boldmath$\uparrow$ \textbf{2.00} & \boldmath$\uparrow$ \textbf{5.00$\times$} & \boldmath$\uparrow$ \textbf{5.69$\times$} &$\uparrow$ 11.93$\times$ &$\uparrow$ 1.00 & \boldmath$\uparrow$ \textbf{5.11$\times$} & \boldmath$\uparrow$ \textbf{6.52$\times$} & \boldmath$\uparrow$ \textbf{8.19$\times$} \\
      & Single   & \boldmath$\uparrow$ \textbf{2.00} &$\uparrow$ 4.82$\times$ &$\uparrow$ 5.55$\times$ &$\uparrow$ 12.33$\times$ &$\uparrow$ 1.00 &$\uparrow$ 3.55$\times$ &$\uparrow$ 4.12$\times$ &$\uparrow$ 8.07$\times$ \\
      & Average  & \boldmath$\uparrow$ \textbf{2.00} &$\uparrow$ 4.86$\times$ &$\uparrow$ 5.60$\times$ &$\uparrow$ 12.56$\times$ & \boldmath$\uparrow$ \textbf{2.00} &$\uparrow$ 4.48$\times$ &$\uparrow$ 5.71$\times$ &$\uparrow$ 8.12$\times$ \\
      & Complete & \boldmath$\uparrow$ \textbf{2.00} &$\uparrow$ 4.85$\times$ &$\uparrow$ 5.59$\times$ &$\uparrow$ 12.30$\times$ &$\downarrow$ 1.00 &$\uparrow$ 2.64$\times$ &$\uparrow$ 2.88$\times$ &$\uparrow$ 7.29$\times$ \\
      & Centroid & \boldmath$\uparrow$ \textbf{2.00} &$\uparrow$ 4.73$\times$ &$\uparrow$ 5.38$\times$ & \boldmath$\uparrow$ \textbf{12.93$\times$} &$\uparrow$ 1.00 &$\uparrow$ 2.81$\times$ &$\uparrow$ 3.12$\times$ &$\uparrow$ 7.11$\times$ \\
      \midrule
      \multirow{5}{*}{$\Delta_{GRAG}$}
      & Ward & \boldmath$\uparrow$ \textbf{1.00} &$\uparrow$ 3.37$\times$ &$\uparrow$ 3.84$\times$ &$\uparrow$ 13.77$\times$ & \boldmath$\downarrow$ \textbf{1.00} & \boldmath$\uparrow$ \textbf{1.39$\times$} & \boldmath$\uparrow$ \textbf{1.50$\times$} &$\uparrow$ 2.78$\times$ \\
      & Single & \boldmath$\uparrow$ \textbf{1.00} &$\uparrow$ 3.51$\times$ &$\uparrow$ 3.98$\times$ &$\uparrow$ 14.00$\times$ &$\downarrow$ 2.00  &$\uparrow$ 1.30$\times$ &$\uparrow$ 1.39$\times$ & \boldmath$\uparrow$ \textbf{2.85$\times$} \\
      & Average  & \boldmath$\uparrow$ \textbf{1.00} &$\uparrow$ 3.61$\times$ & \boldmath$\uparrow$ \textbf{4.08$\times$} & \boldmath$\uparrow$ \textbf{19.77$\times$} &$\downarrow$ 4.00  &$\uparrow$ 1.32$\times$ &$\uparrow$ 1.43$\times$ &$\uparrow$ 2.78$\times$ \\
      & Complete & \boldmath$\uparrow$ \textbf{1.00} &$\uparrow$ 3.60$\times$ & \boldmath$\uparrow$ \textbf{4.08$\times$} &$\uparrow$ 13.41$\times$ & \boldmath$\downarrow$ \textbf{1.00} &$\uparrow$ 1.36$\times$ &$\uparrow$ 1.45$\times$ &$\uparrow$ 2.79$\times$ \\
      & Centroid & \boldmath$\uparrow$ \textbf{1.00} & \boldmath$\uparrow$ \textbf{3.64$\times$} &$\uparrow$ 3.62$\times$ &$\uparrow$ 14.18$\times$ & \boldmath$\downarrow$ \textbf{1.00} &$\uparrow$ 1.29$\times$ &$\uparrow$ 1.39$\times$ &$\uparrow$ 2.78$\times$ \\
      \bottomrule
    \end{tabular}
  }
  \vspace{-3mm}
\end{table}

\begin{table}
  \caption{Effect of different in-batch query size on both datasets (Backbone: Llama-3.2-3B).}
  \label{tab:size_3B}
  \centering
  \setlength{\tabcolsep}{8pt}
  \resizebox{1.0\textwidth}{!}{
    \begin{tabular}{l|cccc|cccc}
      \toprule
      \multirow{2}{*}{Methods} & \multicolumn{4}{c|}{Scene Graph} & \multicolumn{4}{c}{OAG} \\
      & ACC$\uparrow$ & RT$\downarrow$ & TTFT$\downarrow$ & PFTT$\downarrow$ & ACC$\uparrow$ & RT$\downarrow$ & TTFT$\downarrow$ & PFTT$\downarrow$ \\
      \midrule
      \addlinespace[1.0ex]
      \multicolumn{9}{c}{\textbf{50 in-batch queries}} \\
      \midrule
      G-Retriever & 58.00  & 479.90 & 458.56 & 308.45 & 98.00  & 386.50 & 331.04 & 222.13\\
      G-Retriever+SubGCache & \textbf{64.00} & \textbf{155.96} & \textbf{134.94} & \textbf{28.02} & \textbf{100.00} & \textbf{192.98} & \textbf{140.92} & \textbf{33.00} \\
     $\Delta_{G-Retriever}$ &$\uparrow$ 6.00  &$\uparrow$ 3.08$\times$ &$\uparrow$ 3.40$\times$ &$\uparrow$ 11.01$\times$ &$\uparrow$ 2.00  &$\uparrow$ 2.00$\times$ &$\uparrow$ 2.35$\times$ &$\uparrow$ 6.73$\times$ \\
      \midrule
      GRAG & \textbf{58.00} & 260.42 & 251.51 & 396.92 & \textbf{100.00} & 248.94 & 192.28 & 83.41 \\
      GRAG+SubGCache & \textbf{58.00} & \textbf{80.82} & \textbf{68.75} & \textbf{30.10} & \textbf{100.00} & \textbf{181.44} & \textbf{127.00} & \textbf{28.68} \\ 
     $\Delta_{GRAG}$ & 0.00 &$\uparrow$ 3.22$\times$ &$\uparrow$ 3.66$\times$ &$\uparrow$ 13.19$\times$ & 0.00 &$\uparrow$ 1.37$\times$ &$\uparrow$ 1.51$\times$ &$\uparrow$ 2.91$\times$ \\
      \midrule
      \addlinespace[1.0ex]
      \multicolumn{9}{c}{\textbf{150 in-batch queries}} \\
      \midrule
      G-Retriever & 64.00  & 643.15 & 621.96 & 316.34 & \textbf{97.33} & 547.08 & 491.47 & 221.23 \\
      G-Retriever+SubGCache & \textbf{65.33} & \textbf{145.15} & \textbf{123.35} & \textbf{28.07} & \textbf{97.33} & \textbf{184.59} & \textbf{134.36} & \textbf{29.00} \\
     $\Delta_{G-Retriever}$ &$\uparrow$ 1.33  &$\uparrow$ 4.43$\times$ &$\uparrow$ 5.04$\times$ &$\uparrow$ 11.27$\times$ & 0.00  &$\uparrow$ 2.96$\times$ &$\uparrow$ 3.66$\times$ &$\uparrow$ 7.63$\times$ \\
      \midrule
      GRAG &$58.67$ & 543.09 & 786.74 & 400.69 & \textbf{98.67} & 237.65 & 184.10 & 80.52 \\
      GRAG+SubGCache & \textbf{59.33} & \textbf{162.81} & \textbf{206.61} & \textbf{29.76} & \textbf{98.67} & \textbf{179.81} & \textbf{130.32} & \textbf{29.91} \\ 
     $\Delta_{GRAG}$ &$\uparrow$ 0.66  &$\uparrow$ 3.34$\times$ &$\uparrow$ 3.81$\times$ &$\uparrow$ 13.46$\times$ & 0.00 &$\uparrow$ 1.32$\times$ &$\uparrow$ 1.41$\times$ &$\uparrow$ 2.69$\times$ \\
      \midrule
      \addlinespace[1.0ex]
      \multicolumn{9}{c}{\textbf{200 in-batch queries}} \\
      \midrule
      G-Retriever & \textbf{64.50} & 439.39 & 418.35 & 306.75 & 97.00  & 475.00 & 420.67 & 214.68 \\
      G-Retriever+SubGCache & \textbf{64.50} & \textbf{130.14} & \textbf{111.27} & \textbf{25.33} & \textbf{98.00} & \textbf{190.39} & \textbf{139.42} & \textbf{30.70} \\
     $\Delta_{G-Retriever}$ & 0.00  &$\uparrow$ 3.38$\times$ &$\uparrow$ 3.76$\times$ &$\uparrow$ 12.11$\times$ &$\uparrow$ 1.00  &$\uparrow$ 2.49$\times$ &$\uparrow$ 3.02$\times$ &$\uparrow$ 6.99$\times$ \\
      \midrule
      GRAG &$58.00$ & 541.30 & 521.25 & 400.44 & \textbf{99.00} & 249.59 & 192.45 & 80.04 \\
      GRAG+SubGCache & \textbf{60.00} & \textbf{160.60} & \textbf{136.02} & \textbf{28.93} &$98.50$ & \textbf{184.14} & \textbf{132.25} & \textbf{29.04} \\ 
     $\Delta_{GRAG}$ &$\uparrow$ 2.00  &$\uparrow$ 3.37$\times$ &$\uparrow$ 3.83$\times$ &$\uparrow$ 13.84$\times$ &$\downarrow$ 0.50  &$\uparrow$ 1.36$\times$ &$\uparrow$ 1.46$\times$ &$\uparrow$ 2.76$\times$ \\
      \bottomrule
    \end{tabular}
  }
  \vspace{-3mm}
\end{table}

\begin{figure}
\vspace{-2mm}
\centering\includegraphics[width=0.98\textwidth]{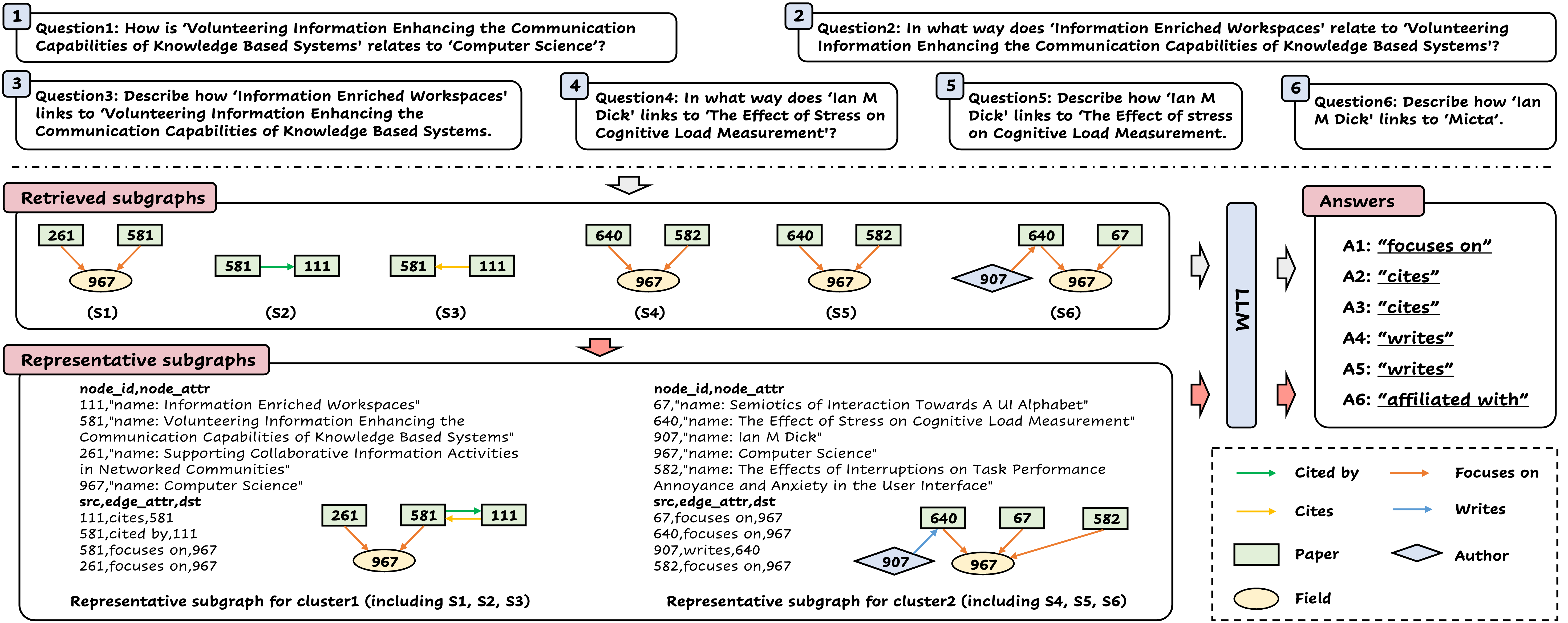}
  \caption{Case study.}
  \label{fig:case_study}
\end{figure}

\subsection{Sensitivity Analysis}
\label{sec:sensitivity}
\noindent \textbf{The choice of linkage strategy.}
To assess the sensitivity of SubGCache to clustering choices, we test five standard linkage strategies: Ward, Single, Average, Complete, and Centroid. As shown in Table~\ref{tab:linkage}, SubGCache consistently achieves substantial latency reduction in RT, TTFT, and PFTT, while maintaining comparable accuracy across all strategies. This confirms that SubGCache is robust and flexible to the clustering methods and performs reliably across diverse linkage strategies.

\noindent \textbf{Impact of in-batch size.} We further evaluate SubGCache under varying in-batch sizes: 50, 100 (from Table~\ref{tab:main}), 150, and 200, using Llama-3.2-3B. The results are reported in Table~\ref{tab:size_3B}, with additional evaluations using Llama-2-7B, Mistral-7B, and Falcon-7B provided in Appendix~\ref{app:inbatch_other_models}. As observed, SubGCache consistently reduces latency while preserving and often improving generation quality across different in-batch sizes. These results demonstrate that SubGCache scales well with in-batch size, supporting its practicality in real-world applications.

\subsection{Case Study}
\label{sec:case}
Figure~\ref{fig:case_study} compares how a batch of example queries is processed with and without SubGCache. Without SubGCache, each query is processed separately using its own retrieved subgraph. In contrast, SubGCache clusters similar queries (\textit{i.e.}, $q_1$–$q_3$ and $q_4$–$q_6$) and constructs a representative subgraph for each cluster, enabling shared KV cache reuse. These representative subgraphs retain all relevant nodes and relations. Both methods generate correct answers, showing that SubGCache significantly accelerates inference without compromising generation quality.  
\section{Conclusion}
\label{sec:conclusion}
This paper introduces a new research problem: in-batch query processing for graph-based RAG, aiming to reduce inference latency through batch-level optimization. To address this, we propose SubGCache, a novel subgraph-level caching framework that tackles the problem-specific challenges of identifying and exploiting structural redundancy in retrieved subgraphs. SubGCache is simple, plug-and-play, and easily integrable into existing graph-based RAG approaches. Experiments across various LLM backbones and graph-based RAG frameworks demonstrate that SubGCache significantly reduces inference latency, while preserving and even improving generation quality.

\newpage
\bibliographystyle{plainnat}
\bibliography{ref}

\newpage
\appendix
\begin{table}
  \caption{Datasets.}
  \label{tab:data_detail}
  \centering
  \setlength{\tabcolsep}{5pt}
  \resizebox{1.00\textwidth}{!}{
  \begin{tabular}{llll}
    \toprule
    Dataset & Textual Graph & Question & Answer \\
    \midrule
    \makecell[l]{Scene\\Graph} & \makecell[l]{node id,node attr\\0,"name: eye glasses; attribute: black; (x,y,w,h): (330, 125, 25, 7)"\\1,"name: laptop; (x,y,w,h): (67, 170, 62, 60)"\\2,"name: cords; attribute: blue; (x,y,w,h): (0, 182, 110, 109)"\\3,"name: windows; (x,y,w,h): (395, 0, 105, 58)"\\4,"name: man; (x,y,w,h): (447, 102, 52, 231)"\\5,"name: woman; (x,y,w,h): (304, 109, 78, 224)"\\6,"name: jeans; (x,y,w,h): (382, 265, 77, 68)"\\7,"name: table; (x,y,w,h): (70, 222, 53, 12)"\\8,"name: man; (x,y,w,h): (370, 108, 58, 205)"\\9,"name: sweater; attribute: orange; (x,y,w,h): (307, 142, 74, 116)"\\10,"name: screen; attribute: on; (x,y,w,h): (0, 78, 90, 111)"\\11,"name: table; attribute: silver; (x,y,w,h): (244, 162, 66, 75)"\\12,"name: windows; attribute: glass; (x,y,w,h): (297, 17, 111, 172)"\\13,"name: pants; attribute: red; (x,y,w,h): (317, 252, 52, 80)"\\14,"name: face; (x,y,w,h): (332, 113, 21, 33)"\\15,"name: shirt; attribute: blue, plaid; (x,y,w,h): (375, 133, 102, 163)"\\16,"name: building; (x,y,w,h): (0, 0, 499, 329)"\\17,"name: eye glasses; (x,y,w,h): (421, 110, 25, 9)"\\18,"name: man; (x,y,w,h): (373, 89, 100, 242)"\\19,"name: man; (x,y,w,h): (117, 53, 143, 280)"\\20,"name: camera; (x,y,w,h): (371, 106, 62, 75)"\\21,"name: suit; attribute: gray; (x,y,w,h): (113, 100, 146, 233)"\\src,edge attr,dst\\0,to the right of,21\\0,to the left of,4\\0,to the left of,8\\0,to the right of,19\\0,to the left of,18\\0,to the left of,20\\1,to the left of,11\\1,to the left of,19\\ \dots} & \makecell[l]{What is the color of the cords?} & \makecell[l]{blue}  \\ \midrule
    OAG & \makecell[l]{node id,node attr\\
0,"name: a dynamic environment for video surveillance"\\1,"name: is the writing on the wall for tabletops"\\2,"name: university of castilla la mancha"\\
3,"name: aalborg university copenhagen"\\
4,"name: queen mary university of london"\\5,"name: panayiotis zaphiris"\\
6,"name: antonietta grasso"\\
7,"name: gilbert cockton"\\8,"name: artificial intelligence"\\
9,"name: computer vision"\\ \dots \\src,edge attr,dst\\
0,written by,963\\
0,focuses on,967\\
1,written by,942\\
1,focuses on,967\\
1,cites,455\\2,has member,895\\
2,has member,896\\
2,has member,897\\ \dots} & \makecell[l]{How is "cross cultural understanding\\ of content and interface in the context\\ of e learning systems" connected to\\ "computer science"?} & focuses on 
    \\
    \bottomrule
  \end{tabular}
  }
\end{table}

\section{Experiments}
\label{sec:exp_detail}

\begin{table}
  \caption{Effect of different in-batch query size on both datasets (Backbone: Llama-2-7B).}
  \label{tab:size_llama7B}
  \centering
  \setlength{\tabcolsep}{9pt}
  \resizebox{1.0\textwidth}{!}{
    \begin{tabular}{l|cccc|cccc}
      \toprule
      \multirow{2}{*}{Model} & \multicolumn{4}{c|}{Scene Graph} & \multicolumn{4}{c}{OAG} \\
      & ACC $\uparrow$ & RT $\downarrow$ & TTFT $\downarrow$ & PFTT $\downarrow$ & ACC $\uparrow$ & RT $\downarrow$ & TTFT $\downarrow$ & PFTT $\downarrow$ \\
      \midrule
      \addlinespace[1.0ex]
      \multicolumn{9}{c}{\textbf{50 in-batch queries}} \\
      \midrule
      G-Retriever & 54.00  & 873.84 & 843.88 & 683.25 & \textbf{96.00} & 838.56 & 716.62 & 482.61 \\
      G-Retriever+SubGCache & \textbf{64.00} & \textbf{180.98} & \textbf{154.04} & \textbf{45.95} & \textbf{96.00} & \textbf{750.40} & \textbf{677.68} & \textbf{61.64} \\
      $\Delta_{G-Retriever}$ & $\uparrow$ 10.00  & $\uparrow$ 4.83$\times$ & $\uparrow$ 5.48$\times$ & $\uparrow$ 14.87$\times$ & 0.00 & $\uparrow$ 1.12$\times$ & $\uparrow$ 1.12$\times$ & $\uparrow$ 7.83$\times$ \\
      \midrule
      GRAG & \textbf{54.00} & 1073.04 & 1041.12 & 923.91 & \textbf{100.00} & 400.52 & 327.40 & 222.16 \\
      GRAG+SubGCache & \textbf{54.00} & \textbf{257.84} & \textbf{223.90} & \textbf{50.44} & 98.00 & \textbf{225.40} & \textbf{187.42} & \textbf{59.22} \\ 
     $\Delta_{GRAG}$ & 0.00 & $\uparrow$ 4.16$\times$ & $\uparrow$ 4.65$\times$ & $\uparrow$ 18.32$\times$ & $\downarrow$ 2.00  & $\uparrow$ 1.57$\times$ & $\uparrow$ 1.75$\times$ & $\uparrow$ 3.75$\times$ \\
      \midrule
      \addlinespace[1.0ex]
      \multicolumn{9}{c}{\textbf{150 in-batch queries}} \\
      \midrule
      G-Retriever & 57.33  & 1345.37 & 1301.90 & 694.75 & \textbf{95.33} & 773.88 & 702.56 & 477.40 \\
      G-Retriever+SubGCache & \textbf{64.00} & \textbf{170.97} & \textbf{142.93} & \textbf{44.62} & \textbf{95.33} & \textbf{641.43} & \textbf{573.56} & \textbf{65.05} \\
      $\Delta_{G-Retriever}$ & $\uparrow$ 6.67  & $\uparrow$ 7.87$\times$ & $\uparrow$ 9.11$\times$ & $\uparrow$ 15.57$\times$ & 0.00 & $\uparrow$ 1.21$\times$ & $\uparrow$ 1.22$\times$ & $\uparrow$ 7.34$\times$ \\
      \midrule
      GRAG & 54.00 & 1199.54 & 1744.63 & 923.17 & 98.67 & 439.55 & 374.89 & 211.83 \\
      GRAG+SubGCache & \textbf{55.33} & \textbf{219.27} & \textbf{281.54} & \textbf{51.63} & \textbf{99.33} & \textbf{247.41} & \textbf{181.47} & \textbf{61.83} \\ 
      $\Delta_{GRAG}$ & $\uparrow$ 1.33  & $\uparrow$ 5.47$\times$ & $\uparrow$ 6.20$\times$ & $\uparrow$ 17.88$\times$ & $\uparrow$ 0.66  & $\uparrow$ 1.78$\times$ & $\uparrow$ 2.07$\times$ & 3.43$\times$ \\
      \midrule
      \addlinespace[1.0ex]
      \multicolumn{9}{c}{\textbf{200 in-batch queries}} \\
      \midrule
      G-Retriever & 58.50  & 827.87 & 796.95 & 676.80 & \textbf{96.50} & 699.36 & 629.36 & 462.83 \\
      G-Retriever+SubGCache & \textbf{66.00} & \textbf{171.57} & \textbf{144.15} & \textbf{46.00} & 94.50  & \textbf{631.47} & \textbf{562.98} & \textbf{62.29} \\
      $\Delta_{G-Retriever}$ & $\uparrow$ 7.50  & $\uparrow$ 4.83$\times$ & $\uparrow$ 5.53$\times$ & $\uparrow$ 14.71$\times$ & $\downarrow$ 2.00  & $\uparrow$ 1.11$\times$ & $\uparrow$ 1.12$\times$ & $\uparrow$ 7.43$\times$ \\
      \midrule
      GRAG & \textbf{54.50} & 1118.74 & 1085.86 & 924.04 & 99.00 & 420.10 & 353.33 & 208.07 \\
      GRAG+SubGCache & \textbf{54.50} & \textbf{216.18} & \textbf{182.19} & \textbf{51.20} & \textbf{99.50} & \textbf{240.28} & \textbf{172.20} & \textbf{61.57} \\ 
      $\Delta_{GRAG}$ & 0.00 & $\uparrow$ 5.18$\times$ & $\uparrow$ 5.96$\times$ & $\uparrow$ 18.05$\times$ & $\uparrow$ 0.50  & $\uparrow$ 1.76$\times$ & $\uparrow$ 2.05$\times$ & 3.38$\times$ \\
      \bottomrule
    \end{tabular}
  }
\end{table}

\subsection{Datasets}
\label{sec:data_detail}
We evaluate SubGCache on two newly constructed datasets: Scene Graph and OAG. Existing GraphQA benchmarks~\cite{he2024g} typically associate each textual graph with a single query, overlooking the in-batch query setting. To bridge this gap, we adapt and construct datasets that support in-batch queries for graph-based RAG. Table~\ref{tab:data_detail} presents the textual graph details and showcases example queries with their answers from both datasets.
\begin{itemize}[leftmargin=*]
    \item \textbf{Scene Graph.} Based on the original Scene Graph dataset from~\cite{he2024g}, we select a graph with 22 nodes and 147 edges, representing objects, attributes, and relationships within an image. We manually construct 426 queries targeting specific entities or relations, with answers grounded in node or edge attributes. Many of these queries require multi-hop reasoning. The dataset is split into 113/113/200 queries for training, validation, and testing, respectively.
    \item \textbf{OAG.} The original OAG~\cite{zhang2019oag,zhu2025hierpromptlm} is a textual graph with various types of nodes and edges. To adapt it to our setting, we construct a query set by sampling 3,434 link prediction queries, where each query involves predicting the relation type between two entities. The dataset is split into 1,617/1,617/200 for training, validation, and testing, respectively.
\end{itemize}

\subsection{Setup}
\label{sec:config}
\noindent \textbf{Baseline models and LLM backbones.} We use G-Retriever~\cite{he2024g} and GRAG~\cite{hu2024grag} as our baselines, and evaluate SubGCache by integrating it as a plug-and-play module during inference, resulting in the variants G-Retriever+SubGCache and GRAG+SubGCache. We primarily adopt Llama-3.2-3B~\cite{grattafiori2024llama} as the backbone LLM, and further test with Llama-2-7B~\cite{touvron2023llama}, Mistral-7B~\cite{mosaicml2023introducing}, and Falcon-7B~\cite{penedo2023refinedweb} to assess SubGCache's scalability and robustness across larger LLMs.

\noindent \textbf{Architecture and Configuration.}
For graph retrieval, we follow the default pipeline of the original baselines~\cite{he2024g,hu2024grag}: SentenceBERT~\cite{reimers2019sentence} is used to encode node and edge attributes, as well as queries, for both G-Retriever and GRAG. For G-Retriever and its SubGCache variant, we select the top-$k$ nodes and edges with $k=3$ and set the edge cost to 0.5. For GRAG and its SubGCache variant, we select the top-$k$ subgraphs with $k=3$ and include the top-$10$ entities within two hops. For graph encoding, G-Retriever and its SubGCache variant use a Graph Transformer~\cite{shi2020masked}, while GRAG and GRAG+SubGCache adopt GAT~\cite{velivckovic2017graph}. Both encoders are configured with 4 layers, 4 attention heads per layer, and a hidden dimension aligned with the LLM backbone. The maximum input sequence length is set to 1024, and the number of generated tokens is capped at 32. For clustering, we adopt agglomerative hierarchical clustering with Euclidean distance and determine cluster assignments by cutting the dendrogram at a predefined number of clusters. 

\noindent \textbf{Training and Evaluation Protocol.}
All models are trained using the AdamW optimizer~\cite{loshchilov2017decoupled} with a learning rate of 1e-5 and weight decay of 0.05. 
Training runs for up to 10 epochs with early stopping: a patience of 2 is used for G-Retriever, and 5 for GRAG. Following both baselines~\cite{he2024g,hu2024grag}, the LLM backbone remains frozen.

During inference, SubGCache is integrated in a plug-and-play manner without modifying any components of the original models. G-Retriever and G-Retriever+SubGCache share the same pretrained G-Retriever model; similarly, GRAG and GRAG+SubGCache share the same pretrained GRAG model. For the main evaluation, we randomly sample 100 test queries from each dataset. All experiments are conducted on two NVIDIA A100-SXM4-40GB GPUs.

\begin{table}
  \caption{Effect of different in-batch query size on both datasets (Backbone: Mistral-7B).}
  \label{tab:size_mistral7B}
  \centering
  \setlength{\tabcolsep}{9pt}
  \resizebox{1.0\textwidth}{!}{
    \begin{tabular}{l|cccc|cccc}
      \toprule
      \multirow{2}{*}{Model} & \multicolumn{4}{c|}{Scene Graph} & \multicolumn{4}{c}{OAG} \\
      & ACC $\uparrow$ & RT $\downarrow$ & TTFT $\downarrow$ & PFTT $\downarrow$ & ACC $\uparrow$ & RT $\downarrow$ & TTFT $\downarrow$ & PFTT $\downarrow$ \\
      \midrule
      \addlinespace[1.0ex]
      \multicolumn{9}{c}{\textbf{50 in-batch queries}} \\
      \midrule
      G-Retriever & \textbf{66.00} & 838.20 & 807.58 & 716.72 & \textbf{100.00} & 729.71 & 646.97 & 511.57 \\
      G-Retriever+SubGCache & \textbf{66.00} & \textbf{210.76} & \textbf{181.06} & \textbf{49.38} & \textbf{100.00} & \textbf{317.86} & \textbf{229.59} & \textbf{62.46} \\
      $\Delta_{G-Retriever}$ & 0.00 & $\uparrow $3.98$\times$ & $\uparrow $4.46$\times$ & $\uparrow$ 14.51$\times$ & 0.00 & $\uparrow$ 2.30$\times$ & $\uparrow$ 2.82$\times$ & $\uparrow$8.19$ \times$ \\
      \midrule
      GRAG & 58.00 & 1113.50 & 1082.32 & 967.28 & \textbf{100.00} & 440.92 & 358.78 & 248.16 \\
      GRAG+SubGCache & \textbf{62.00} & \textbf{227.32} & \textbf{195.08} & \textbf{18.80} & \textbf{100.00} & \textbf{237.78} & \textbf{159.26} & \textbf{60.43} \\ 
      $\Delta_{GRAG}$ & $\uparrow$ 4.00  & $\uparrow$ 4.90$\times$ & $\uparrow$ 5.55$\times$ & $\uparrow$ 18.80$\times$ & 0.00 & $\uparrow$ 1.85$\times$ & $\uparrow$ 2.25$\times$ & 4.11$\times$ \\
      \midrule
      \addlinespace[1.0ex]
      \multicolumn{9}{c}{\textbf{150 in-batch queries}} \\
      \midrule
      G-Retriever & \textbf{68.00} & 1336.57 & 1290.55 & 729.47 & \textbf{99.33} & 755.09 & 677.22 & 503.59 \\
      G-Retriever+SubGCache & \textbf{68.00} & \textbf{414.58} & \textbf{384.83} & \textbf{50.00} & \textbf{99.33} & \textbf{303.50} & \textbf{228.99} & \textbf{64.13} \\
      $\Delta_{G-Retriever}$ & 0.00 & $\uparrow$ 3.22$\times$ & $\uparrow$ 3.35$\times$ & $\uparrow$ 14.59$\times$ & 0.00 & $\uparrow$ 2.49$\times$ & $\uparrow$ 2.96$\times$ & $\uparrow$ 7.85$\times$ \\
      \midrule
      GRAG & 57.33 & 1114.03 & 1623.39 & 966.18 & \textbf{99.33} & 456.81 & 379.37 & 233.84 \\
      GRAG+SubGCache & \textbf{65.33} & \textbf{193.37} & \textbf{243.95} & \textbf{52.84} & \textbf{99.33} & \textbf{241.03} & \textbf{165.21} & \textbf{65.43} \\ 
      $\Delta_{GRAG}$ & $\uparrow$ 8.00  & $\uparrow$ 5.76$\times$ & $\uparrow$ 6.65$\times$ & $\uparrow$ 18.29$\times$ & 0.00 & $\uparrow$ 1.90$\times$ & $\uparrow$ 2.30$\times$ & 3.57$\times$ \\
      \midrule
      \addlinespace[1.0ex]
      \multicolumn{9}{c}{\textbf{200 in-batch queries}} \\
      \midrule
      G-Retriever & \textbf{68.00} & 865.71 & 834.86 & 712.71 & \textbf{99.50} & 722.10 & 644.00 & 489.87 \\
      G-Retriever+SubGCache & 67.00  & \textbf{350.64} & \textbf{321.06} & \textbf{49.57} & \textbf{99.50} & \textbf{288.64} & \textbf{212.21} & \textbf{63.24} \\
      $\Delta_{G-Retriever}$ & $\downarrow$ 1.00  & $\uparrow$ 2.47$\times$ & $\uparrow$ .60$\times$ & $\uparrow$ 4.38$\times$ & 0.00 & $\uparrow$ 2.50$\times$ & $\uparrow$ 3.03$\times$ & $\uparrow$ 7.75$\times$ \\
      \midrule
      GRAG & 54.50 & 1113.72 & 1081.63 & 966.82 & \textbf{99.50} & 442.55 & 361.28 & 232.05 \\
      GRAG+SubGCache & \textbf{65.00} & \textbf{199.61} & \textbf{169.05} & \textbf{18.52} & \textbf{99.50} & \textbf{244.99} & \textbf{167.18} & \textbf{63.28} \\ 
      $\Delta_{GRAG}$ & $\uparrow$ 10.50  & $\uparrow$ 5.58$\times$ & $\uparrow$ 6.40$\times$ & $\uparrow$ 18.52$\times$ & 0.00 & $\uparrow$ 1.81$\times$ & $\uparrow$ 2.16$\times$ & 3.67$\times$ \\
      \bottomrule
    \end{tabular}
  }
\end{table}

\begin{table}
  \caption{Effect of different in-batch query size on both datasets (Backbone: Falcon-7B).}
  \label{tab:size_falcon7B}
  \centering
  \setlength{\tabcolsep}{9pt}
  \resizebox{1.0\textwidth}{!}{
    \begin{tabular}{l|cccc|cccc}
      \toprule
      \multirow{2}{*}{Model} & \multicolumn{4}{c|}{Scene Graph} & \multicolumn{4}{c}{OAG} \\
      & ACC $\uparrow$ & RT $\downarrow$ & TTFT $\downarrow$ & PFTT $\downarrow$ & ACC $\uparrow$ & RT $\downarrow$ & TTFT $\downarrow$ & PFTT $\downarrow$ \\
      \midrule
      \addlinespace[1.0ex]
      \multicolumn{9}{c}{\textbf{50 in-batch queries}} \\
      \midrule
      G-Retriever & \textbf{62.00} & 913.80 & 879.24 & 672.87 & \textbf{100.00} & 709.12 & 628.12 & 471.46 \\
      G-Retriever+SubGCache & \textbf{62.00} & \textbf{289.08} & \textbf{253.38} & \textbf{53.47} & \textbf{100.00} & \textbf{258.96} & \textbf{176.90} & \textbf{56.43} \\
      $\Delta_{G-Retriever}$ & 0.00 & $\uparrow$ 3.16$\times$ & $\uparrow$ 3.47$\times$ & $\uparrow$ 12.58$\times$ & 0.00 & $\uparrow$ 2.74$\times$ & $\uparrow$ 3.55$\times$ & $\uparrow$ 8.35$\times$ \\
      \midrule
      GRAG & \textbf{56.00} & 1101.98 & 1065.00 & 953.93 & \textbf{100.00} & 433.24 & 346.60 & 201.20 \\
      GRAG+SubGCache & \textbf{56.00} & \textbf{243.54} & \textbf{209.48} & \textbf{54.06} & \textbf{100.00} & \textbf{273.54} & \textbf{191.42} & \textbf{57.75} \\ 
      $\Delta_{GRAG}$ & 0.00 & $\uparrow$ 4.52$\times$ & $\uparrow$ 5.08$\times$ & $\uparrow$ 17.65$\times$ & 0.00 & $\uparrow$ 1.58$\times$ & $\uparrow$ 1.81$\times$ & 3.48$\times$ \\
      \midrule
      \addlinespace[1.0ex]
      \multicolumn{9}{c}{\textbf{150 in-batch queries}} \\
      \midrule
      G-Retriever & 65.33  & 1027.01 & 1311.02 & 684.68 & \textbf{98.00} & 710.75 & 633.24 & 472.24 \\
      G-Retriever+SubGCache & \textbf{69.33} & \textbf{208.33} & \textbf{174.15} & \textbf{51.03} & 97.33  & \textbf{253.10} & \textbf{173.18} & \textbf{59.34} \\
      $\Delta_{G-Retriever}$ & $\uparrow$ 4.00  & $\uparrow$ 4.93$\times$ & $\uparrow$ 7.53$\times$ & $\uparrow$ 13.42$\times$ & $\downarrow$ 0.67  & $\uparrow$ 2.81$\times$ & $\uparrow$ 3.66$\times$ & $\uparrow$ 7.96$\times$ \\
      \midrule
      GRAG & 56.67 & 1122.63 & 1628.99 & 956.46 & \textbf{97.33} & 473.41 & 391.93 & 193.60 \\
      GRAG+SubGCache & \textbf{60.00} & \textbf{250.13} & \textbf{324.87} & \textbf{52.19} & \textbf{97.33} & \textbf{258.30} & \textbf{179.96} & \textbf{60.87} \\ 
      $\Delta_{GRAG}$ & $\uparrow$ 3.33  & $\uparrow$ 4.49$\times$ & $\uparrow$ 5.01$\times$ & $\uparrow$ 18.33$\times$ & 0.00 & $\uparrow$ 1.83$\times$ & $\uparrow$ 2.18$\times$ & 3.18$\times$ \\
      \midrule
      \addlinespace[1.0ex]
      \multicolumn{9}{c}{\textbf{200 in-batch queries}} \\
      \midrule
      G-Retriever & 65.50  & 825.75 & 789.16 & 669.30 & \textbf{98.50} & 687.32 & 607.52 & 459.11 \\
      G-Retriever+SubGCache & \textbf{68.50} & \textbf{186.71} & \textbf{153.70} & \textbf{49.89} & \textbf{98.50} & \textbf{626.26} & \textbf{544.64} & \textbf{61.53} \\
      $\Delta_{G-Retriever}$ & $\uparrow$ 3.00  & $\uparrow$ 4.42$\times$ & $\uparrow$ 5.13$\times$ & $\uparrow$ 13.42$\times$ & 0.00 & $\uparrow$ 1.10$\times$ & $\uparrow$ 1.12$\times$ & $\uparrow$ 7.46$\times$ \\
      \midrule
      GRAG & 57.50 & 1121.91 & 1082.84 & 958.31 & \textbf{98.00} & 454.89 & 371.87 & 192.12 \\
      GRAG+SubGCache & \textbf{59.50} & \textbf{226.08} & \textbf{192.64} & \textbf{53.22} & \textbf{98.00} & \textbf{235.13} & \textbf{155.99} & \textbf{56.96} \\ 
      $\Delta_{GRAG}$ & $\uparrow$ 2.00  & $\uparrow$ 4.96$\times$ & $\uparrow$ 5.62$\times$ & $\uparrow$ 18.01$\times$ & 0.00 & $\uparrow$ 1.93$\times$ & $\uparrow$ 2.38$\times$ & 3.37$\times$ \\
      \bottomrule
    \end{tabular}
  }
\end{table}

\subsection{Metrics}  
\label{sec:metrics}
We evaluate all models in the in-batch query setting using four key metrics that jointly assess generation quality and inference efficiency:
\begin{itemize}[leftmargin=*]
\item \textbf{Accuracy (ACC).}
ACC measures the proportion of correctly answered queries, serving as the primary metric for generation quality. 

\item \textbf{Response Time (RT).}  
RT denotes the total end-to-end latency for each query, measured from the moment the query is submitted to the completion of the full model response. This includes subgraph retrieval, prompt construction, LLM prefill, and token generation.

\item \textbf{Time to First Token (TTFT).}  
TTFT measures the time from query submission to the generation of the first output token. It reflects the system's responsiveness, which is especially important in latency-sensitive applications.

\item \textbf{Prefill and First Token Time (PFTT).}  
PFTT isolates the portion of TTFT that corresponds to the LLM's prefill computation and first-token generation. It directly reflects the effectiveness of KV cache reuse and prompt reuse strategies.

\end{itemize}

\subsection{Evaluation Across Different In-batch Query Sizes with Other LLM Backbones}
\label{app:inbatch_other_models}
To further assess the scalability of SubGCache across different LLM backbones, we conduct additional experiments using Llama-2-7B, Mistral-7B, and Falcon-7B. Following the same setup as in Table~\ref{tab:size_3B}, we test SubGCache with 50, 100 (from Table~\ref{tab:main}), 150 and 200 in-batch queries on both datasets. The results are presented in Table~\ref{tab:size_llama7B}, Table~\ref{tab:size_mistral7B} and Table~\ref{tab:size_falcon7B}, respectively.  
Consistent trends are observed across different models and in-batch sizes: SubGCache significantly reduces inference latency while maintaining or even improving generation quality. These observations further validate its generalizability and effectiveness across different LLM backbones and in-batch settings, consistent with the findings discussed in Section~\ref{sec:sensitivity}.

\section{Impact Statements}
\label{sec:impact}
This paper aims to advance the field of graph-based RAG systems by addressing a critical gap in inference efficiency improvement. We introduce a new in-batch query setting and explicitly tackle the structural redundancy present in retrieved subgraphs. To this end, we propose SubGCache, the first subgraph-level KV caching framework tailored for graph-based RAG, which significantly reduces inference latency without compromising generation quality. The framework is lightweight, plug-and-play, and model-agnostic, making it easily applicable to a wide range of graph-based RAG systems. We believe this work will inspire further research on caching and batch-level optimization for structure-level generation, offering broad societal benefits without foreseeable negative impacts.

\section{Limitations and Future Work}
\label{sec:limit}
Our current evaluation focuses on specific question-answering (QA) tasks~\cite{siriwardhana2023improving,schick2023toolformer,asai2023self}. In future work, we plan to extend SubGCache to abstract QA settings~\cite{edge2024local,guo2024lightrag}. While currently applied during the inference stage, SubGCache could also be explored during training to further improve efficiency or alignment. These directions are orthogonal to our core contribution and do not diminish its novelty, which lies in pioneering subgraph-level KV cache reuse for efficient graph-based RAG inference.

\end{document}